\pdfoutput=1
\documentclass[11pt]{article}

\usepackage[final]{acl}

\usepackage{times}
\usepackage{latexsym}

\usepackage[T1]{fontenc}
\usepackage[utf8]{inputenc}

\usepackage{microtype}

\usepackage{inconsolata}

\usepackage{graphicx}
\usepackage{arydshln}
\usepackage{algorithm}
\usepackage{algpseudocode}
\usepackage{array} 
\usepackage{amssymb}
\usepackage{booktabs}
\usepackage{array,makecell}
\usepackage{caption}
\usepackage{wrapfig}
\usepackage{enumitem}
\usepackage{amsmath}
\usepackage{bm}
\usepackage{booktabs}
\usepackage{multirow}
\usepackage{rotating}
\usepackage{subcaption}
\usepackage{diagbox}

\title{HiDe-LLaVA: Hierarchical Decoupling for Continual Instruction Tuning of Multimodal Large Language Model}

\author{
 \textbf{Haiyang Guo\textsuperscript{1,2}\thanks{Equal Contribution.}},
 \textbf{Fanhu Zeng\textsuperscript{2,3}\footnotemark[1]},
 \textbf{Ziwei Xiang\textsuperscript{2,3}},
 \textbf{Fei Zhu\textsuperscript{4}},\\
 \textbf{Da-Han Wang\textsuperscript{5}},
 \textbf{Xu-Yao Zhang\textsuperscript{1,2,3}\thanks{Corresponding Author.}},
 \textbf{Cheng-Lin Liu\textsuperscript{1,2,3}}\\ 
 {\normalsize \textsuperscript{1}School of Advanced Interdisciplinary Sciences, UCAS } 
 {\normalsize \textsuperscript{2}MAIS, CASIA} \\
 {\normalsize \textsuperscript{3}School of Artificial Intelligence, UCAS} 
 {\normalsize \textsuperscript{4}Centre for Artificial Intelligence and Robotics, HKISI-CAS } \\
 {\normalsize \textsuperscript{5}FKLPRIU, School of Computer and Information Engineering, Xiamen University of Technology} \\
 {\normalsize \{guohaiyang2023, zengfanhu2022, zhufei2018\}@ia.ac.cn, \{xyz, liucl\}@nlpr.ia.ac.cn}
}

\begin{document}
\maketitle
\begin{abstract}
Instruction tuning is widely used to improve a pre-trained Multimodal Large Language Model~(MLLM) by training it on curated task-specific datasets, enabling better comprehension of human instructions. However, it is infeasible to collect all possible instruction datasets simultaneously in real-world scenarios. Thus, enabling MLLM with continual instruction tuning is essential for maintaining their adaptability. However, existing methods often trade off memory efficiency for performance gains, significantly compromising overall efficiency. In this paper, we propose a task-specific expansion and task-general fusion framework based on the variations in Centered Kernel Alignment~(CKA) similarity across different model layers when trained on diverse datasets. Furthermore, we analyze the information leakage present in the existing benchmark and propose a new and more challenging benchmark to rationally evaluate the performance of different methods. Comprehensive experiments showcase a significant performance improvement of our method compared to existing state-of-the-art methods. Code and dataset are released at \url{https://github.com/Ghy0501/HiDe-LLaVA}.
\end{abstract}

\section{Introduction}

Recent years have witnessed remarkable advancements in Multimodal Large Language Models~(MLLMs)~\cite{yin2023survey}, which extend the capabilities of Large Language Models~\cite{touvron2023llama} through sophisticated vision-text feature alignment mechanisms~\cite{liu2024visual} and autoregressive generation frameworks. The integration of large-scale training corpora and extensive model parameters~\cite{yang2024law,kaplan2020scaling} has enabled MLLMs to achieve state-of-the-art performance across diverse downstream applications~\cite{zhao2025chartedit,lu2024mathvista}, demonstrating the potential for complex world understanding and representing a significant milestone toward the realization of artificial general intelligence.

As a pivotal component for MLLMs, instruction tuning~\cite{zhang2023instruction} enhances instruction-following capabilities of pre-trained models, effectively bridging general-purpose pretraining and domain-specific applications by aligning model behavior with user intent. However, in practical applications, users often perform continuous fine-tuning on diverse datasets at different times to meet specific needs. This requires the model to effectively incorporate new knowledge while overcoming catastrophic forgetting~\cite{li2017learning, kirkpatrick2017overcoming} on previous tasks.

Recent work~\cite{chen2024coin} construct a benchmark to evaluate the capability of MLLMs in continual instruction tuning~(CIT) and reveal that there is a serious catastrophic forgetting phenomenon in MLLMs. In terms of methodology, they propose MoELoRA to mitigate the model's forgetting of old instructions. Based on this, ModalPrompt~\cite{zeng2024modalprompt} dynamically selects optimal prompts during inference by jointly leveraging textual and visual features, thereby enhancing model performance. However, we observe that the downstream datasets used by these methods partially overlaps with the tasks encountered during the supervised fine-tuning~(SFT) phase of MLLM. Such information leakage~\cite{kim2023learnability} compromises the reliability of evaluating whether a method mitigates forgetting or if the model inherently retains this capability, thereby diminishing the challenge of continual instruction tuning.

In this paper, we first select the instruction tuning datasets that the model has not encountered during the SFT phase by evaluating the model's zero-shot performance on specific task. Therefore, our reconstructed continual instruction tuning dataset prioritizes instructions that are unfamiliar or unknown to the model, enabling a more accurate and fair comparison of the performance across different methods. To tackle the issue of catastrophic forgetting, we first investigate the CKA similarity~\cite{kornblith2019similarity} of the model's outputs between the same layers on different instruction tuning datasets and observe that the model exhibits a significant similarity difference between the top layer and the remaining layers. Based on this observation, our experimental analysis indicates that the model focuses more on task-specific information in the top layer, while learning more generic knowledge in the remain layers. Therefore, we propose a \textbf{Hi}erarchical \textbf{De}coupling method named \textbf{HiDe-LLaVA}, which consists of two simple yet effective strategies: task-general fusion and task-specific expansion. Comprehensive results verified that our method effectively overcoming the catastrophic forgetting during continual instruction tuning. 

In summary, our main contributions include: 
\begin{itemize}
    \item We propose HiDe-LLaVA, which enhances the continual instruction tuning performance by decoupling the model layers into task-specific expansion and task-general fusion.
    \item We reveal information leakage in existing benchmark and propose a more fair continual instruction tuning benchmark to equitably assess the effects of different methods.
    \item Extensive experimental results show that our method effectively overcomes the catastrophic forgetting during continual instruction tuning.
\end{itemize}

\section{Relate Work}

\textbf{Multimodal Large Language Models.}
With the predominant capability of large language models~(LLMs)~\cite{touvron2023llama,zhang2023llama} in handing natural language processing tasks~\cite{wang2022progress,min2023recent}, huge amount of efforts has been made to multimodal learning and multimodal large language models~(MLLMs)~\cite{bai2023qwen,zhu2024minigpt,dai2024instructblip} are proposed to tackle task in multimodal scenarios. In addition to a large language model, which is composed of stacks of transformer blocks~\cite{vaswani2017attention}, MLLMs also incorporates a vision encoder~\cite{dosovitskiy2020vit} and a feature fusion module to align the cross-modality representations~(usually a projection layer~\cite{liu2024visual} or cross attention~\cite{dai2024instructblip}). Answers are then generated in response to multimodal inputs. The paradigm is of great significance and achieves impressive performance on various downstream tasks~\cite{liu2025mmbench,zhao2025chartcoder,zhang2024mm}.

\noindent
\textbf{Continual Instruction Tuning.}
Instruction tuning~\cite{ouyang2022training, zhang2023instruction} aims to enpower MLLMs to better understand and follow human instructions. To adapt to dynamically changing instructions in real-world scenarios, MLLMs need the capability to learn current and retain previous learned instructions, which is named \emph{continual instruction tuning}. In recent studies, Eproj~\cite{he2023continual} first constructs a benchmark in continual instruction tuning settings and reveal that catastrophic forgetting is still observed in MLLMs. However, the benchmark they proposed is insufficient in terms of the number and diversity of tasks and is not evaluated on mainstream MLLM architectures. By contrast, CoIN~\cite{chen2024coin} establishes a diversity benchmark consisting of 8 crafted datasets and proposed MoELoRA to mitigate catastrophic forgetting on various MLLMs. More recently, ModalPrompt~\cite{zeng2024modalprompt} further enhances performance in this setting by leveraging image and text modalities to guide the selection of appropriate prompts from the prompt pool. Continual-LLaVA~\cite{cao2024continual} similarly proposes a pool of LoRAs~\cite{hu2021lora} to select the appropriate LoRAs for training based on textual similarity, and using them during inference.

\begin{figure*}[t]
    \centering
    \begin{subfigure}[b]{0.65\textwidth}
    \centering
    \includegraphics[width=1.0\textwidth, height=0.33\textwidth]{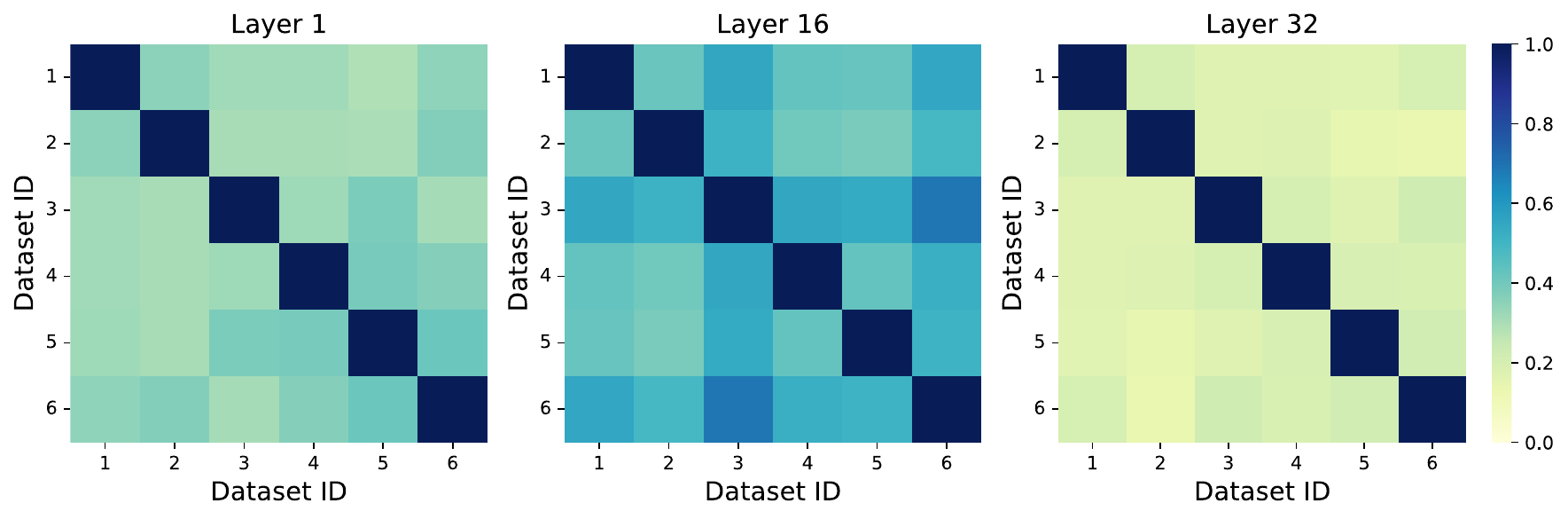}
    \vspace{-1.5em}
    \label{fig:CKA_heatmap}
  \end{subfigure}%
  \hfill
  \begin{subfigure}[b]{0.35\textwidth}
    \centering
    \includegraphics[width=1.0\textwidth, height=0.60\textwidth]{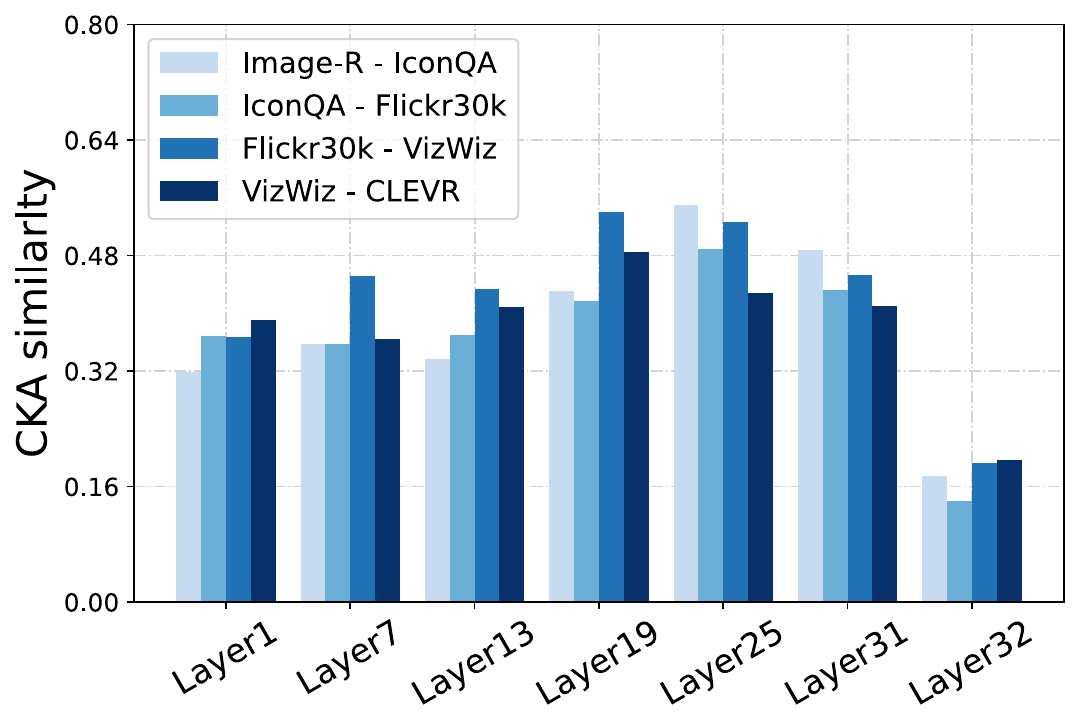}
    \vspace{-1.5em}
    \label{fig:CKA_bar}
  \end{subfigure}%
  \vspace{-5pt}
  \caption{Left: Output CKA similarity heatmaps for different task inputs across the bottom, middle, and top layers. Overall, the output similarity across different tasks markedly decreases at the top layer. Right: Detailed similarity comparison of different task pairs. It can be seen that even for these very different pairs~(IconQA and Flickr30k), the similarity differences only appear in last layers and most layers are similar.}
  \label{fig:CKA_sim}
\vspace{-15pt}
\end{figure*}

\section{Preliminary}
\label{preliminary}
Continual instruction tuning for MLLMs aims to address the problem of learning with continuous data and mitigate catastrophic forgetting. 
Assume that there are $T$ tasks in total, and the MLLM parameterized by $\theta$ has been pre-trained on large-scale image-text data to obtain aligned multimodal features. The sequential tasks data of continual instruction tuning can be expressed as: $\mathcal{D}_t = \{ (\textbf{x}_{v}^{t,j}, \textbf{x}_{ins}^{t,j}, \textbf{x}_{ans}^{t,j})_{j=1}^{N_t}\}$, $t\in \{ 1,...,T\}$, where $\textbf{x}_{v}^{t,j}, \textbf{x}_{ins}^{t,j}$ and $\textbf{x}_{ans}^{t,j}$ denote the input image tokens, instruction tokens and answer tokens, $N_t$ is the total number of image-text pairs of task $t$. Taking a simple image-text pair with an answer in length $L$, the objective of MLLM is to predict next token based on all the preceding tokens in an autoregressive way:
\begin{equation}
    \mathcal{L}_{MLLM}^{t} = -\sum_{l=1}^{L} \log p(\textbf{x}_{ans}^{l} | \textbf{x}_{v}^{t}, \textbf{x}_{ins}^{t}, \textbf{x}_{ans}^{<l}; \theta).
\label{eq:mllm}
\end{equation}

When learning task $t$, the goal of continual instruction tuning is to maximize the retention of knowledge from the learned task while preserving the model's generalization ability for unseen tasks.

\section{Methodology}
\label{sec:methodology}

\subsection{Hierarchical decoupling of MLLM}
\label{CKA}

Existing research~\cite{kornblith2019similarity} shows that the outputs between different layers of the model exhibit notable differences, indicating that the model focuses on different patterns of the input at each layer. Inspired by this, we fine-tune LLaVA-v1.5~\cite{liu2024visual} using LoRA on 6 instruction tuning datasets and analyze the output CKA similarity~\cite{kornblith2019similarity} differences across the same layers between models.\footnote{The details of the CKA similarity computation are presented in Appendix~\ref{sec:cka_cal}.} As illustrated in left of Fig~\ref{fig:CKA_sim}, the similarity of the model outputs exhibits a clear difference between the top and remaining layers. Specifically, for two datasets that differ significantly in both input image styles and textual instructions~(IconQA~\cite{lu2021iconqa} and Flickr30k~\cite{flickrentitiesijcv}), we observe that their output similarity differs notably only in the top layer~(\emph{i.e.} the layer closest to the output), while remaining relatively high across the other layers. This suggests that for different tasks, the model primarily learns similar, generalized knowledge in the layers below the top layer, while the top layer focuses more on dissimilar, task-specific information. Therefore, we hypothesize that for continual instruction tuning, the shared information can be obtained by integrating the layers beneath the top layer, while task-specific knowledge in the top layer should be selectively activated.

\begin{figure}[t]
    \centering
    \includegraphics[width=0.98\linewidth]{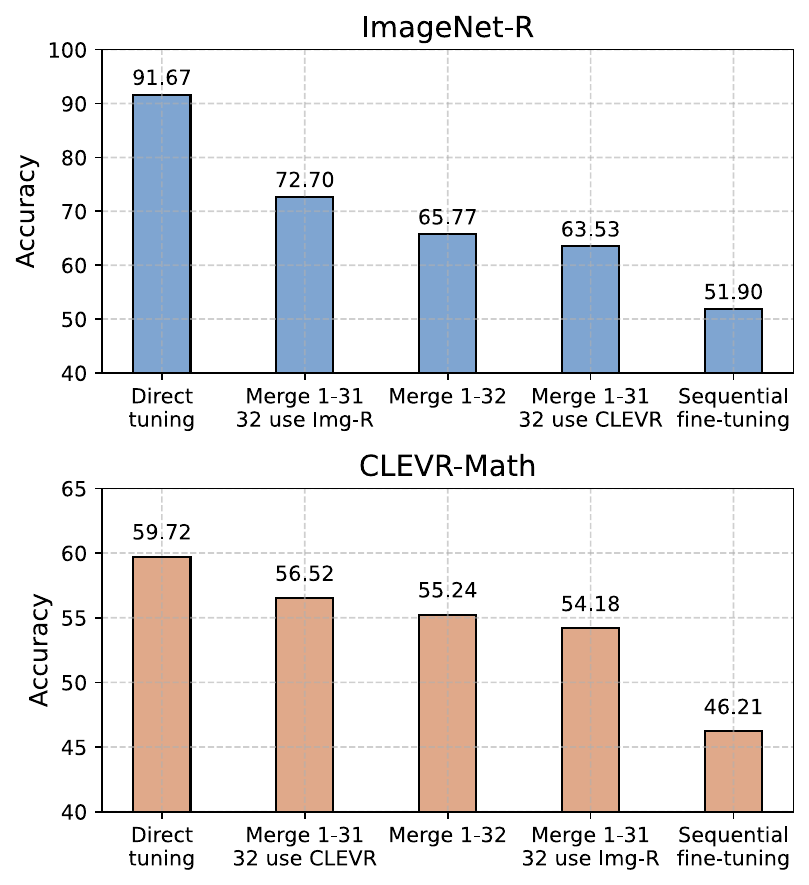}
    \vspace{-5pt}
    \caption{Impact of different LoRA operational strategies on individual task performance.}
    \label{fig:toy-exp}
    \vspace{-20pt}
\end{figure}

\begin{figure*}[t]
    \centering
    \includegraphics[width=0.98\linewidth]{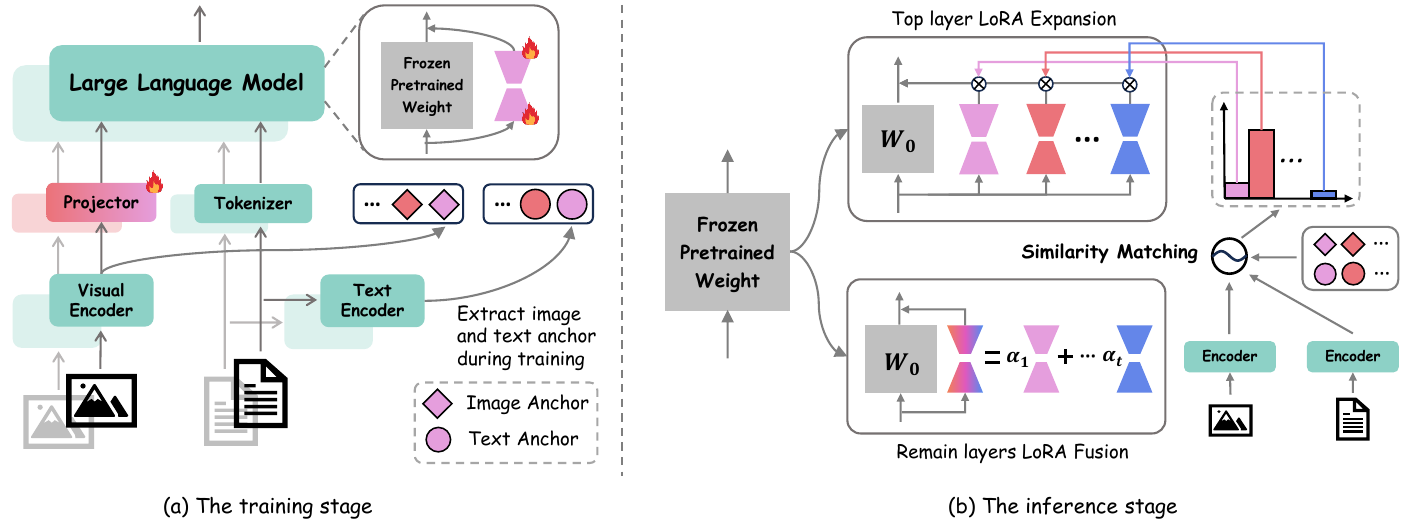}
    \vspace{-10pt}
    \caption{An overview of HiDe-LLaVA framework. (a) During training, we optimize the LoRA modules and projector layer with an autoregressive loss and the image-text anchors are extracted from the image and text encoders of CLIP. (b) At inference time, our method apply a MoE-like expansion on the top-layer LoRA and dynamically distribute expert weights via similarity matching with previously learned image and text anchors. For the remaining layers, general knowledge across tasks is incorporated through LoRA fusion.}
    \label{fig:framework}
\vspace{-15pt}
\end{figure*}

To further validate our hypothesis, we performe LoRA fine-tuning on two instruction tuning datasets, ImageNet-R~\cite{hendrycks2021many} and CLEVR-Math~\cite{lindstrom2022clevr}, to obtain their respective LoRA weights. We then apply the following operations to the LoRAs at different layers: \textbf{(\emph{i})} applying corresponding LoRAs to all layers, \textbf{(\emph{ii})} applying corresponding LoRAs only to the top layer~(32-th layer) while merging the rest~(1-31 layers), \textbf{(\emph{iii})} merging LoRAs across all layers, \textbf{(\emph{iv})} applying only the top layer with non-corresponding LoRAs while merging the rest of the layers, and \textbf{(\emph{v})} fine-tuning the same set of LoRAs in sequence. As shown in Fig~\ref{fig:toy-exp}, directly merging all LoRA layers for two tasks provides greater mitigation of catastrophic forgetting compared to sequential fine-tuning~(\emph{i.e.}~\emph{iii} and \emph{v}), but also leads to performance degradation for the respective tasks. However, merging only the layers below the top layer and correctly selecting task-specific LoRAs for the top layer results in further performance improvements. By contrast, selecting non-corresponding LoRAs for the top layer can degrade the model's performance on the corresponding task~(\emph{i.e.}~\emph{ii} and \emph{iv}).

Building on the above analysis, we posit that catastrophic forgetting can be effectively mitigated through hierarchical decoupling, integrating task-general knowledge across tasks, and adaptively selecting task-specific information. The framework of our method is shown in Fig.~\ref{fig:framework}.

\vspace{-5pt}
\subsection{Proposed Method: HiDe-LLaVA}
Following the LoRA fine-tuning strategy used in LLaVA~\cite{liu2024improved}, we embed LoRA modules into all linear layers of the language model. During training, LoRA modules and the projector layer are trained to align with the current input instructions. For clarity, we use $E$ to represent all LoRA modules in the following.

\subsubsection{Task-specific Expansion on Top Layer}
As mentioned above, the model emphasizes task-specific knowledge at the top layer, making it important to allocate the output appropriately. Therefore, adaptively selecting the appropriate LoRA module based on the inputs, without relying on a Task-ID, becomes a critical challenge. Inspired by prototype learning~\cite{zhu2021prototype, tan2022fedproto}, we propose extracting high-dimensional feature representations of input data during training as an alternative to explicit Task-ID assignment. Specifically, the image features can be extracted directly by the LLaVA's visual encoder, while on the text side, we introduce CLIP’s text encoder~\cite{radford2021learning} to derive feature representations from the input instructions:
\begin{equation}
    m_{v}^{t} = \frac{1}{N_{t}}\sum_{n=1}^{N_t}f_{v}(\textbf{x}_{v}^{t}),~m_{ins}^{t} = \frac{1}{N_{t}}\sum_{n=1}^{N_t}f_{ins}(\textbf{x}_{ins}^{t}),
\label{eq:extract}
\end{equation}
\noindent
where $\textbf{x}_{v}^{t}$ and $\textbf{x}_{\text{ins}}^{t}$ denote the input image and instruction for task $t$, respectively, while $m_{v}^{t}$ and $m_{\text{ins}}^{t}$ represent the extracted image and text features, which we collectively refer to as image and text anchors. Here, $f_{v}$ and $f_{\text{ins}}$ correspond to the CLIP image encoder and text encoder, respectively.

With image and text anchors for each task, we expand the top-level LoRA of all learned tasks during inference in a manner similar to a Mixture-of-Experts (MoE)~\cite{shazeer2017outrageously} model. Instead of using a traditional MoE Router, we select the appropriate LoRA's output based on the cosine similarity between the input test data and the anchors. Specifically, we first compute the cosine similarity between the features of the current test input $\textbf{z}^{test}$ and each task anchor:
\begin{equation}
    r_{v}^{c} = \text{sim}(\textbf{z}_{v}^{test}, m_{v}^{c}),~r_{ins}^{c} = \text{sim}(\textbf{z}_{ins}^{test}, m_{ins}^{c}),
\end{equation}
\noindent
where $\text{sim}(\textbf{a}, \textbf{b}) = \frac{\textbf{a} \cdot \textbf{b}}{\| \textbf{a}\|~\| \textbf{b}\|}.$ The similarities of the image and text are then obtained as a normalized score $d_c$ using the softmax function:
\begin{equation}
    d_c = \frac{e^{\bar{r^c} / T}}{\sum_{j=1}^{T} e^{\bar{r^j} / T}},~\text{where}~\bar{r^j} = \alpha \cdot r_{v}^{c} + \beta \cdot r_{ins}^{c},
\label{eq:confidence}
\end{equation}
\noindent
where $T$ denotes the temperature coefficient of Softmax, $\alpha$ and $\beta$ are the hyper-parameters that control the fusion of image and text similarity. The resulting $d_c$ represents the degree of match between the current test input and a particular learned task $c$. We use this in place of the expert weights computed via the router in traditional MoE models, multiplying the output of each top-level LoRA by the corresponding score and summing them to obtain the final output of the entire LoRA branch on the top layer:
\begin{equation}
    O_{top} = \sum_{i=1}^{T} d_i E_{i}(h),
\label{eq:lora_branch}
\end{equation}
\noindent
where $E_i$ denotes the $i$-th LoRA expert and $h$ represents the hidden input of the linear layer. Compared to MoELoRA, our method mitigates catastrophic forgetting by eliminating the need to train a router at each stage, leading to a more effective and stable distribution of LoRA outputs.

\subsubsection{Task-general Fusion on Remain Layers}
\label{sec:fusion}
For the remaining layers outside the top layer, our CKA similarity analysis in Sec~\ref{CKA} reveals a relatively high degree of output similarity, suggesting that these layers encode more generalized knowledge shared across tasks. To integrate this generalized knowledge, we employ a simple and effective model parameter fusion strategy~\cite{ilharco2022editing, zheng2025learning, zeng2025parameter}:
\begin{equation}
    \bar{E}_{T} = \sum_{i=1}^{T} \epsilon_i E_i,
\label{eq:lora_merge}
\end{equation}
\noindent
where $T$ is the number of learned tasks and $\epsilon_i$ denotes the fusion coefficient of $i$-th LoRA module. The output of the remaining layer LoRA branches can then be represented as:
\begin{equation}
    O_{rem} = \bar{E}_{T}(h).
\end{equation}

\begin{algorithm}[t]
    \renewcommand{\algorithmicrequire}{\textbf{Input:}}
    \caption{Pipeline of HiDe-LLaVA}
    \label{alg:framework}
    \begin{algorithmic}[1]
        \Require
        $\mathcal{D}_t : \{ (\textbf{x}_{v}^{t,j}, \textbf{x}_{ins}^{t,j}, \textbf{x}_{ans}^{t,j})_{j=1}^{N_t}\}, t\in \{ 1,..,T\}$. 
        LoRA modules $E$ and Projector layer \emph{Proj}.
        \For{$t=1\to~T$}
            \State \textcolor{gray}{\# Training stage.}
            \For{$(\textbf{x}_{v}^{t,j}, \textbf{x}_{ins}^{t,j}, \textbf{x}_{ans}^{t,j}) \in~\mathcal{D}_t$}
                \State Train $E_t$ and \emph{Proj} through Eq~(\ref{eq:mllm}).
                \State Extract anchors through Eq~(\ref{eq:extract}).
            \EndFor
            \State \textcolor{gray}{\# Inference stage.}
            \For{$\mathcal{D}_t^{test}$ $\in$ $\{\mathcal{D}_1^{test},..., \mathcal{D}_T^{test}\}$}
                \State \textcolor{gray}{\# Task-specific Expansion.}
                \State Compute $d_c$ through Eq~(\ref{eq:confidence}).
                \For{$E$ embedded in top layer}
                    \State \parbox[t]{\dimexpr\linewidth-6em}{Expand the LoRA branch in a MoE manner and compute the output through Eq~(\ref{eq:lora_branch}).}
                \EndFor
                \State \textcolor{gray}{\# Task-general Fusion.}
                \For{$E$ embedded in remain layers}
                    \State \parbox[t]{\dimexpr\linewidth-6em}{Fuse the LoRA modules of all learned tasks through Eq~(\ref{eq:lora_merge}).}
                \EndFor
            \EndFor
        \EndFor
    \end{algorithmic}
\end{algorithm}

In summary, our proposed HiDe-LLaVA hierarchically decomposes the model into a top-level LoRA extension and the fusion of LoRAs in the remaining layers. The overall framework is summarized in Algorithm~\ref{alg:framework}.

\section{Experiments}
\label{sec:experiments}

\begin{table*}[t]
    \centering
    \resizebox{0.95\linewidth}{!}{
    \begin{tabular}{c>{\raggedright\arraybackslash}p{2.7cm} >{\centering\arraybackslash}p{1.5cm} >{\centering\arraybackslash}p{1.5cm}>{\centering\arraybackslash}p{1.5cm} >{\centering\arraybackslash}p{1.5cm} >{\centering\arraybackslash}p{1.5cm} >{\centering\arraybackslash}p{1.5cm} >{\centering\arraybackslash}p{2cm}}
        \toprule
           & {\hspace{1em}} Method & \textbf{Image-R} & \textbf{ArxivQA} & \textbf{Viz-cap} & \textbf{IconQA} & \textbf{CLEVR} & \textbf{Flickr30k} & {{\textbf{Average}}} \\
  
        \midrule

            \multirow{3}{*}{ }& {\hspace{1em}}Zero-shot & 16.27 & 53.73 & 38.39 & 19.20 & 20.63 & 41.88 & 31.68   \\   
            \multirow{3}{*}{ }& {\hspace{1em}}Multi-task & 90.63 & 91.30 & 61.81 & 73.90 & 73.60 & 57.45 & 74.78   \\

            \midrule\midrule
            \multirow{7}{*}{\rotatebox{90}{\textbf{Avg}}} &{\hspace{1em}}FineTune & 49.31 & 78.40 & 50.48 & 53.44 & 55.53 &  \underline{57.95} & 57.52  \\
            & {\hspace{1em}}LwF  & 55.60 & 79.86 & 53.23 & 54.87 & 56.51 & 56.34 & 59.40  \\
            & {\hspace{1em}}EWC  & 54.23 & 80.13 & 53.14 & 55.06 & \underline{57.52} & 55.94 & 59.34  \\
            & {\hspace{1em}}L2P  & 41.52 & 82.32 & 51.98 & 52.21 & 43.16 & 52.77 & 53.99  \\
            & {\hspace{1em}}O-LoRA & \underline{75.26} & \underline{86.73} & \textbf{55.86} & \underline{58.47} & 57.38 & 53.52 & \underline{64.54}  \\
            & {\hspace{1em}}MoELoRA  & 64.49 & 82.42 & 49.54 & 56.87 & 56.35 & \textbf{58.34} & 61.33  \\
            & {\hspace{1em}}\textbf{HiDe-LLaVA} &\textbf{85.70}&\textbf{92.70}&\underline{54.10}&\textbf{66.87}&\textbf{59.12}&55.15&\textbf{68.94}~\textcolor[rgb]{0.81,0,0}{\textbf{(+4.4)}}\\
            \midrule
            % ----------------------------------------------------
            \multirow{7}{*}{\rotatebox{90}{\textbf{Last}}} &{\hspace{1em}}FineTune & 37.63 & 72.33 & 43.47 & 41.7 & 35.63 & \underline{57.95} & 48.12  \\
            & {\hspace{1em}}LwF  & 40.27 & 75.93 & 42.76 & 44.38 & 37.43 & 56.34 & 49.52  \\
            & {\hspace{1em}}EWC  & 39.05 & 77.88 & 43.24 & 45.33 & 39.72 & 55.94 & 50.20  \\
            & {\hspace{1em}}L2P  & 32.73 & 80.41 & 43.72 & 42.16 & 39.25 & 52.77 & 48.51  \\
            & {\hspace{1em}}O-LoRA & \underline{69.36} & \underline{82.42} & \underline{48.64} & \underline{53.66} & \underline{42.53} & 53.52 & 58.36  \\
            & {\hspace{1em}}MoELoRA  & 49.87 & 77.63 & 43.65 & 46.40 & 36.47 & \textbf{58.34} & 52.06  \\
            & {\hspace{1em}}\textbf{HiDe-LLaVA}& \textbf{80.50}&\textbf{89.83}&\textbf{48.78}&\textbf{62.90}&\textbf{47.97}&55.15&\textbf{64.19}~\textcolor[rgb]{0.81,0,0}{\textbf{(+5.8)}}\\
        \bottomrule
    \end{tabular}}
    \vspace{-5pt}
    \caption{Comparison with various methods on our UCIT benchmark in terms of \emph{Avg} and \emph{Last}. The best and second methods are labeled with \textbf{bold} and \underline{underline} styles.  Our method outperforms the best previous methods by \textbf{4.4\%} and \textbf{5.8\%} in Avg and Last metrics, respectively.}
    \label{tab:ucit_table}
     \vspace{-15pt}
\end{table*}

\subsection{Experimental Setup}
\noindent
\textbf{Datasets}\quad We train and evaluate the effectiveness of our method on (1) CoIN~\cite{chen2024coin} benhcmark and (2) our reconstructed benchmark. In particular, CoIN consists of datasets such as VQAv2~\cite{goyal2017making}, VizWiz~\cite{gurari2018vizwiz}, ScienceQA~\cite{lu2022learn}, TextVQA~\cite{singh2019towards}, GQA~\cite{hudson2019gqa}, OCR-VQA~\cite{mishra2019ocr}, ImageNet~\cite{deng2009imagenet} and REC-COCO~\cite{kazemzadeh2014referitgame, mao2016generation}. A major limitation of CoIN is the overlap between the selected downstream 
datasets and the dataset used during LLaVA's pre-training phase, resulting in information leakage~\cite{kim2023learnability} that significantly undermines fair comparisons between different methods. To address this, we screen six datasets that are highly uncorrelated with the pre-training data based on LLaVA's zero-shot performance~(See Zero-shot in Table~\ref{tab:ucit_table}), specifically: ArxivQA~\cite{li2024multimodal}, CLEVR-Math~\cite{lindstrom2022clevr}, IconQA~\cite{lu2021iconqa}, ImageNet-R~\cite{hendrycks2021many}, VizWiz-caption~\cite{gurari2018vizwiz} and Flickr30k~\cite{plummer2015flickr30k}. We term this the \textbf{U}nseen \textbf{C}ontinual \textbf{I}nstruction \textbf{T}uning~(\textbf{UCIT}) benchmark. Details are provided in Appendix~\ref{app:detail} and~\ref{app:visualization}.

\noindent
\textbf{Evaluation Metrics.} Following the standard metrics in conrinual learning~\cite{wang2022learning}. We report \emph{Last}, \emph{Avg} to evaluate the continual instruction tuning performance. \emph{Last} is computed as the accuracy of all seen tasks after learning the final task and \emph{Avg} is the average accuracy of each task during the training process. These two metrics measure the model's capacity to retain learned tasks.

\noindent
\textbf{Baseline.} We compare our HiDe-LLaVA with traditional continual learning methods and recent continual instruction tuning method. For the former, we implement LwF~\cite{li2017learning}, EWC~\cite{kirkpatrick2017overcoming}, L2P~\cite{wang2022learning} and O-LoRA~\cite{wang2023orthogonal} within a MLLM architecture, meticulously tuning parameters to ensure reliable and effective results. For the latter, We compare with MoELoRA~\cite{chen2024coin} to highlight the superiority of our method. The performance of zero-shot and multi-task fine-tuning is also reported to represent the lower and upper bounds for each benchmark. Specific details of each method can be found in Appendix~\ref{sec:comparison}.

\noindent
\textbf{Implementation details.} We use LLaVA-v1.5-7b~\cite{liu2024visual} as the base multimodal model and CLIP-L/14-336~\cite{radford2021learning} to extract visual and textual features. Following LLaVA's LoRA fine-tuning strategy, we embed LoRA modules in all linear layers of the language model with the rank set to 8. The temperature coefficient $T$ and hyper-parameters in Eq~(\ref{eq:confidence}) are set to 0.1, 0.5 and 0.5, respectively. The fusion coefficient in Eq~(\ref{eq:lora_merge}) are uniformly set to 1.0. We set the training epoch for all tasks to 1 and the warm-up ratio to 0.03.The learning rates for LoRA and the projector are set to 2e-4 and 2e-5, respectively, with a cosine decay schedule. We set the batch size to 24 for all methods and run the experiments on A800 GPUs.

\begin{figure}[t]
    \centering
    \includegraphics[width=1.0\linewidth]{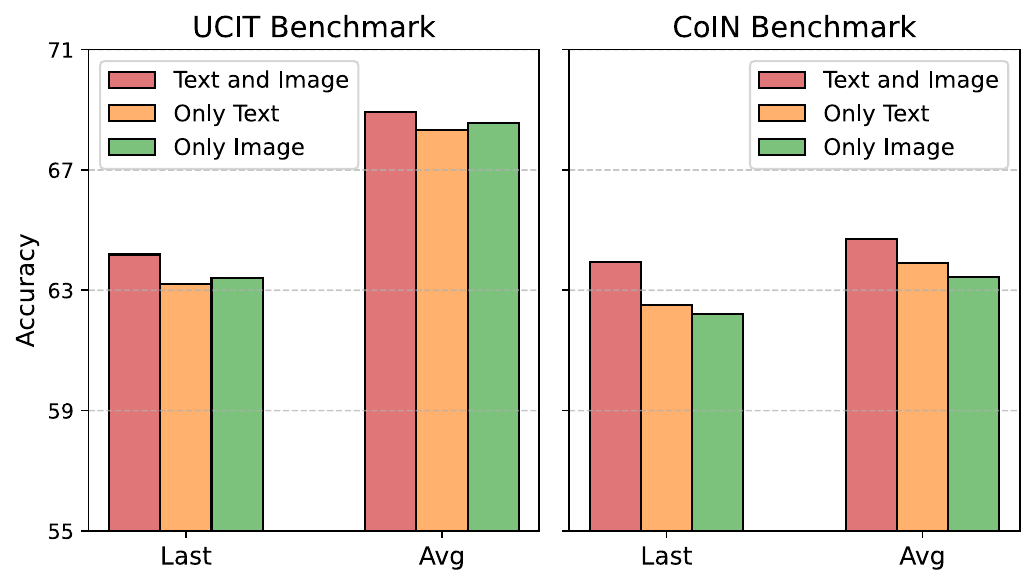}
    \caption{Ablation studies of dual-modalities similarity matching on UCIT and CoIN benchmark.}
    \vspace{-15pt}
    \label{fig:dual-modalities}
\end{figure}

\begin{table*}[t]
    \centering
    \resizebox{0.95\linewidth}{!}{
    \begin{tabular}{c>{\raggedright\arraybackslash}p{3cm} >{\centering\arraybackslash}p{1cm} >{\centering\arraybackslash}p{1cm}>{\centering\arraybackslash}p{1cm} >{\centering\arraybackslash}p{1cm} >{\centering\arraybackslash}p{1cm}>{\centering\arraybackslash}p{1cm} >{\centering\arraybackslash}p{1cm} >{\centering\arraybackslash}p{1cm} >{\centering\arraybackslash}p{2cm}}
        \toprule
           & {\hspace{1em}} Method & \textbf{SciQA} & \textbf{Image} & \textbf{Viz} & \textbf{REC} & \textbf{Text} & \textbf{GQA} & \textbf{VQA} & \textbf{OCR} & {{\textbf{Average}}} \\
  
        \midrule
        
            \multirow{3}{*}{ }& {\hspace{1em}}Zero-shot & 69.79 & 9.93 & 45.50 & 58.47 & 57.75 & 60.77 & 66.50 & 64.93 & 54.21   \\
            \multirow{3}{*}{ }& {\hspace{1em}}Multi-task & 82.36 & 89.63 & 52.51 & 65.83 & 61.27 & 59.93 & 65.67 & 62.03 & 67.40   \\

            \midrule \midrule
            \multirow{7}{*}{\rotatebox{90}{\textbf{Avg}}} &{\hspace{1em}}FineTune  & 64.22 & 40.13 & 43.87 & 38.32 & 55.04 & 55.89 & 60.61 & 64.78 & 52.86  \\
            & {\hspace{1em}}LwF  & 65.20 & 40.63 & 43.22 & 40.05 & 56.23 & 54.67 & 60.64 & 65.12 & 53.22  \\
            & {\hspace{1em}}EWC  & 65.11 & 40.89 & 44.09 & 39.67 &  54.92 & 56.03 & 61.12 & 64.55 & 53.30  \\
            & {\hspace{1em}}L2P  & 70.52 & 26.89 & 45.53 & 45.21 &  56.84 & 59.03 & 63.52 & 64.11 & 53.96  \\
            & {\hspace{1em}}O-LoRA & \underline{73.32} & \underline{68.37} & \underline{50.26} & \underline{61.12} & \textbf{57.75} & \underline{60.96} & \underline{65.71} & 63.31 & \underline{62.60}  \\
            & {\hspace{1em}}MoELoRA  & 68.38 & 48.50 & 44.22 & 40.23 & 55.62 & 57.04 & 62.14 & \textbf{65.75} & 55.24  \\
            & {\hspace{1em}}\textbf{HiDe-LLaVA} &\textbf{74.92}&\textbf{76.72}&\textbf{51.24}&\textbf{61.84}&\underline{57.13}&\textbf{62.83}&\textbf{68.15}&\underline{64.76}&\textbf{64.70}~\textcolor[rgb]{0.81,0,0}{\textbf{(+2.1)}}\\
            \midrule
            % ----------------------------------------------------
            \multirow{7}{*}{\rotatebox{90}{\textbf{Last}}} &{\hspace{1em}}FineTune & 57.43 & 28.90 & 41.88 & 30.05 & 51.39 & 50.76 & 53.28 & 64.78 & 47.31  \\
            & {\hspace{1em}}LwF  & 60.71 & 30.58 & 41.49 & 36.01 & 52.80 & 47.07 & 53.43 & 65.12 & 48.40  \\
            & {\hspace{1em}}EWC  & 59.75 & 31.88 & 42.26 & 34.96 & 51.06 & 51.84 & 55.30 & 64.55 & 48.95  \\
            & {\hspace{1em}}L2P  & 70.21 & 23.31 & 44.21 & 43.76 & 56.25 & 58.46 & 62.32 & 64.11 & 52.83  \\
            & {\hspace{1em}}O-LoRA & \underline{72.56} & \underline{62.84} & \underline{48.43} & \underline{58.97} & \textbf{57.66} & \underline{59.14} & \underline{63.21} & 63.31 & \underline{60.77}  \\
            & {\hspace{1em}}MoELoRA  & 62.02 & 37.21 & 43.32 & 33.22 & 52.05 & 53.12 & 57.92 & \textbf{65.75} & 50.58  \\
            & {\hspace{1em}}\textbf{HiDe-LLaVA}& \textbf{73.20}&\textbf{69.28}&\textbf{50.76}&\textbf{59.18}&\underline{56.92}&\textbf{61.33}&\textbf{67.12}&\underline{64.76}&\textbf{63.95}~\textcolor[rgb]{0.81,0,0}{\textbf{(+3.2)}}\\
        \bottomrule
    \end{tabular}}
    \vspace{-4pt}
    \caption{Comparison with various methods on our CoIN benchmark in terms of \emph{Avg} and \emph{Last}. The best and second methods are labeled with \textbf{bold} and \underline{underline} styles. Our HiDe-LLaVA outperforms the best previous methods by \textbf{2.1\%} and \textbf{3.2\%} in Avg and Last metrics, respectively.}
    \label{tab:coin_table}
     \vspace{-5pt}
\end{table*}

\subsection{Main Results}
Results on UCIT and CoIN benchmark are shown in Table~\ref{tab:ucit_table} and~\ref{tab:coin_table}. Our method achieves state-of-the-art performance on both benchmarks, effectively mitigating catastrophic forgetting in MLLM during continual instruction tuning. Specifically, MoELoRA alleviates forgetting by transforming LoRA fine-tuning into a Mixture-of-Experts~(MoE) model, while simultaneously training a router to assign the appropriate task outputs. Traditional continual learning methods, such as LwF, EWC, and L2P, show limited success, whereas O-LoRA outperforms all other baseline methods in the comparison. However, these methods still exhibit a significant gap when compared to multi-task learning. In contrast, our method not only significantly outperforms other methods in terms of Avg and Last metrics~(with average improvements of \textbf{4.4\%}, \textbf{5.8\%} on the UCIT benchmark, and \textbf{2.1\%}, \textbf{3.2\%} on the CoIN benchmark), but also leads substantially in training and inference efficiency~(See Sec~\ref{sec:further}).

Notably, by comparing the Zero-shot and Multi-task performance on the UCIT and CoIN benchmarks in Table~\ref{tab:ucit_table} and~\ref{tab:coin_table}, we observe serious information leakage in CoIN. Specifically, the average performance difference between Multi-task and Zero-shot in CoIN is only 13.19\%, while in our UCIT benchmark, the difference is significantly higher at \textbf{43.10\%}. Additionally, some datasets in CoIN even performed worse than Zero-shot after fine-tuning~(\emph{e.g.,} GQA, OCRVQA and VQAv2), suggesting that LLaVA may have encountered similar tasks during pre-training. The continued reuse of these datasets could lead to overfitting, thereby impacting the fairness of comparisons between different methods. Additional experimental results are provided in Appendix~\ref{app:further}.

\subsection{Ablation Study}
Our HiDe-LLaVA comprises two main components: the expansion of task-specific knowledge in the top layer and the fusion of task-general information in the remaining layers. We explore their individual impacts on the results in this section.

\begin{table}[t]
	\centering
        \resizebox{\linewidth}{!}{
	\begin{tabular}{lcccccc}
		\toprule		
            Methods &  \textbf{Last} & $\Delta$ &\textbf{Avg}&  $\Delta$ &  \textbf{Param.~(M)} &$\Delta$ \\
		\midrule \midrule
            HiDe-LLaVA & 64.19 & 0.0 & 68.94 & 0.0 & 44.27 & $\times 1$ \\
            Merge (1-32) & 61.26 & \textcolor[rgb]{0, 0, 0.81}{-2.93} & 65.43 & \textcolor[rgb]{0, 0, 0.81}{-3.51} & 38.27 & \textcolor[rgb]{0.81,0,0}{$\times 0.86$} \\
            Merge (1-31) & 60.64 & \textcolor[rgb]{0, 0, 0.81}{-3.55} & 63.28 & \textcolor[rgb]{0, 0, 0.81}{-5.66} & 44.27 &  \textcolor[rgb]{0.81,0,0}{$\times 0.84$} \\
            Expand (1-32) & 67.62 & \textcolor[rgb]{0.81,0,0}{+3.43} & 70.91 & \textcolor[rgb]{0.81,0,0}{+1.97} & 229.62 & \textcolor[rgb]{0, 0, 0.81}{$\times 5.20$} \\
            Expand (1-31) & 66.84 & \textcolor[rgb]{0.81,0,0}{+2.65} & 70.13 & \textcolor[rgb]{0.81,0,0}{+1.19} & 222.42 & \textcolor[rgb]{0, 0, 0.81}{$\times 5.02$} \\
		\bottomrule
	\end{tabular}}
	\caption{Ablation studies of different fusion and expansion strategies conducted on the UCIT benchmark.}
        \vspace{-15pt}
	\label{tab:strategies}
\end{table}

\noindent
\textbf{The effect of dual-modality similarity matching.} At the top layer, we compute the expert weights for each LoRA output by matching the input features of the two modalities, image and text, with the mean values of the features extracted during training. In Fig~\ref{fig:dual-modalities}, we compare the results of calculating similarity using only a single modality and observe that the results for individual modalities are consistently lower than those for dual-modality matching. This is primarily because a single modality cannot fully capture the differences between tasks. For instance, the image inputs for some tasks may all come from the same dataset, yet the textual instructions differ significantly. In contrast, some tasks may all involve question-answering, but the input images vary greatly in style. Therefore, considering both image and text similarities is crucial.

\begin{figure}[t]
    \centering
    \vspace{-5pt}
    \includegraphics[width=1.0\linewidth]{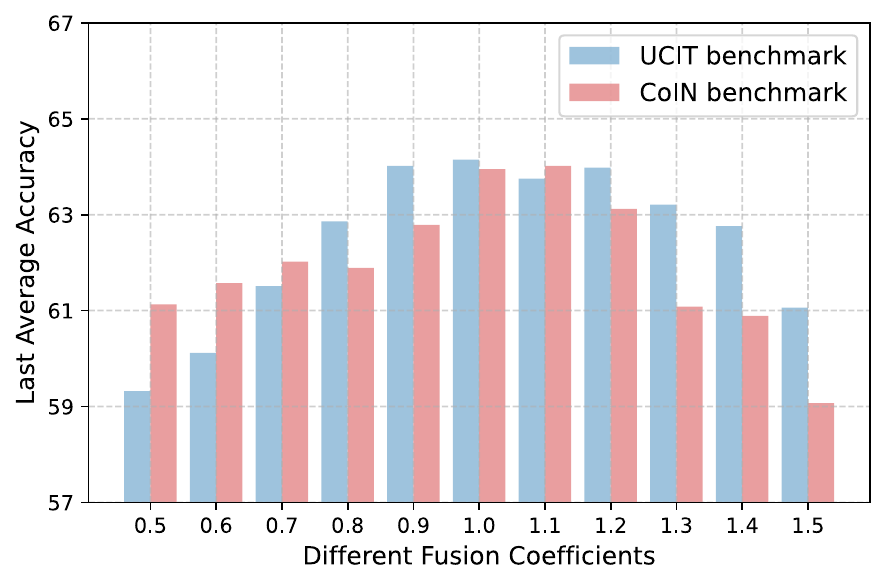}
    \vspace{-20pt}
    \caption{Ablation studies of the fusion coefficient on UCIT and CoIN benchmark.}
    \vspace{-15pt}
    \label{fig:fusion-coefficient}
\end{figure}

\noindent
\textbf{The impact of fusion and expansion strategies.}
Based on the CKA similarity analysis presented in Sec~\ref{CKA}, our HiDe-LLaVA expands LoRA at the top layer and fuse it in the remaining layers. In this section, we compare our method with the results of expanding or fusing LoRA modules across all layers. As shown in Table~\ref{tab:strategies}, the model's performance significantly degrades when all layers are fused, or when all layers except the top layer are fused with LoRA. Although expanding the top layer operation to all layers or the remaining layers yields some performance improvement, it leads to a considerable increase in parameters during inference~(\emph{e.g.,} $\times \textbf{5.20}$), which is unreasonable for long-term continual instruction tuning tasks. Our HiDe-LLaVA, therefore, strikes an optimal balance between performance and parameter efficiency.

\noindent
\textbf{The impact of fusion coefficient.} In order to integrate generalized knowledge across different tasks, we propose a simple yet effective model fusion strategy in Sec~\ref{sec:fusion}, where the choice of parameter fusion coefficients plays a crucial role in the results. In Fig~\ref{fig:fusion-coefficient}, we present a comprehensive results of the impact of different fusion coefficients. Overall, larger fusion coefficients contribute positively to the model’s performance, with a coefficient of 1.0 yielding the best average results across both benchmarks.

\begin{table}[t]
	\centering
        \resizebox{0.9\linewidth}{!}{
	\begin{tabular}{c | >{\centering\arraybackslash}p{1.3cm} >{\centering\arraybackslash}p{1.3cm}>{\centering\arraybackslash}p{1.3cm}}
		\toprule	\diagbox{Metrics}{Order} & \texttt{\textbf{RAVICF}} & \texttt{\textbf{AIRFCV}} &  \texttt{\textbf{IFRCAV}} \\
		\midrule \midrule
            Last & 64.19 & 63.56 & 64.87 \\
            Avg & 68.94 & 68.02 & 69.68 \\
		\bottomrule
	\end{tabular}}
	\caption{Results of different task orders on UCIT benchmark. We adopt an abbreviation scheme to simplify the representation of task sequence notation. For example, \texttt{\textbf{RAVICF}} represents the order of \texttt{ImgNet-\textbf{R}} $\rightarrow$ \texttt{\textbf{A}rxivQA} $\rightarrow$ \texttt{\textbf{V}izWiz-cap} $\rightarrow$ \texttt{\textbf{I}conQA} $\rightarrow$ \texttt{\textbf{C}LEVR-Math} $\rightarrow$ \texttt{\textbf{F}lickr30k}.}
	\label{tab:order}
 \vspace{-15pt}
\end{table}

\subsection{Further Analysis}
\label{sec:further}
\noindent
\textbf{Analysis of Different task order.} To evaluate the robustness of our method across diverse scenarios, we conduct additional experiments with different task sequences on UCIT benchmark. The results, as presented in Table~\ref{tab:order}, demonstrate that our proposed HiDe-LLaVA maintains consistent and stable performance, regardless of the order in which the tasks are presented. This suggests that our method is resilient to task ordering and can effectively handle variations in task sequences during continuous instruction fine-tuning.

\begin{figure}[t]
    \centering
    \includegraphics[width=1.0\linewidth]{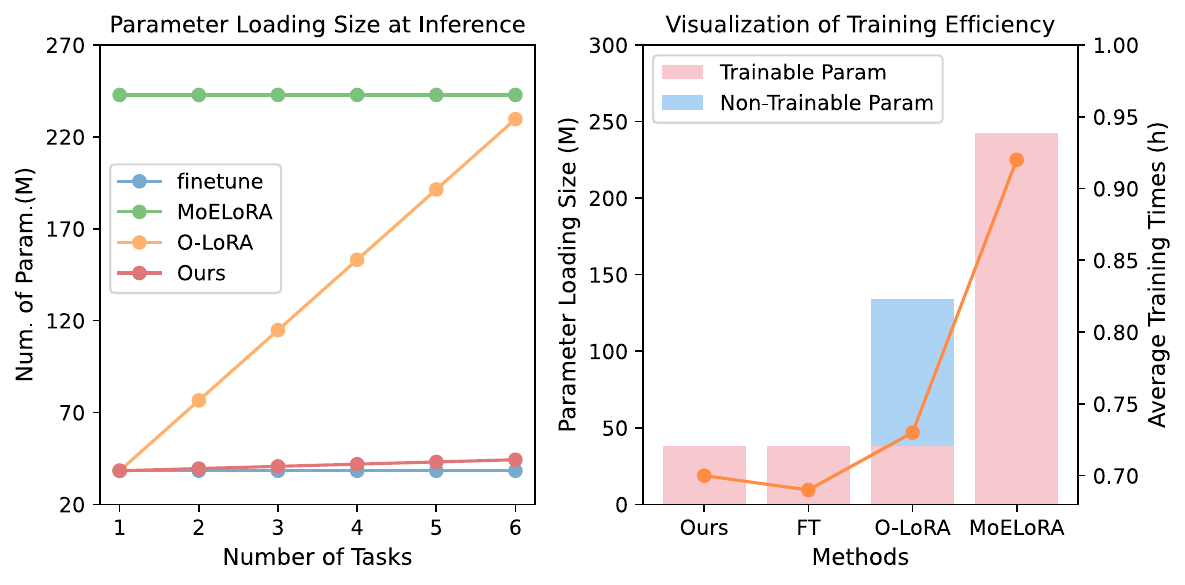}
    \caption{Efficiency analysis on UCIT benchmark. Left: Comparison of the number of parameters loaded during inference. Right: Comparison of training time and loaded parameters.}
    \vspace{-15pt}
    \label{fig:efficiency}
\end{figure}

\noindent
\textbf{Analysis of the efficiency of the training and inference phases.} In Fig.~\ref{fig:efficiency}, we compare the efficiency in terms of both training and inference times across different methods. Specifically, MoELoRA requires the most memory occupied at inference time, as it transforms all embedded LoRAs into the form of Mixture-of-Experts and necessitates prior knowledge of the number of tasks to learn. This limitation is problematic in real-world applications, where predicting the number of future tasks is often infeasible. O-LoRA, on the other hand, concatenates the LoRAs of all learned tasks with the current LoRA, resulting in a substantial increase in memory usage as the number of tasks increases. In contrast, our method only retains the LoRAs of all learned tasks at the top level, which significantly reduces the memory overhead associated with additional parameters.

As for the overhead during training, MoELoRA introduces a large number of trainable parameters, leading to a \textbf{20\%} increase in training time compared to standard fine-tuning. O-LoRA, on the other hand, need to compute the orthogonal loss between the parameter space of the current task and those of previously learned tasks, requiring the inclusion of LoRAs for all past tasks during training. In contrast, our method achieves optimal performance while maintaining training times and efficiency nearly on par with fine-tuning, further demonstrating the effectiveness of our approach.

\section{Conclusion}
In this paper, we investigate how to equip MLLM with the ability to continuously follow user input instructions. Our method first decouples the MLLM into task-specific knowledge layer and task-general knowledge layer based on the differences in CKA similarity of the outputs at each layer between tasks. Then, a task-specific expansion and task-general fusion framework is proposed to mitigate catastrophic forgetting during continual instruction tuning. Additionally, we identify and analyze the information leakage in existing benchmark and introduce more challenging benchmark to ensure a fair evaluation of different methods. Extensive experiments demonstrate that our method not only achieves optimal performance but also maintains high efficiency, making it a well-balanced solution.

\section*{Limitations}
In this work, our proposed HiDe-LLaVA introduces a post-processing framework designed for task-general knowledge fusion and task-specific knowledge expansion. While it achieves state-of-the-art performance, the framework is constrained by performance degradation caused by model fusion operations. We argue that investigating methods to reduce inter-task conflicts during training could further alleviate the issue of catastrophic forgetting. 

Moreover, while this paper and existing research primarily address the forgetting phenomenon of old tasks in the context of continual instruction tuning, we emphasize the importance of preserving the original capabilities of large multimodal models like LLaVA, which exhibit strong zero-shot generalization abilities to unseen tasks. Ensuring that such models retain their foundational strengths under continuous instruction fine-tuning remains a critical challenge, and we plan to explore this direction in future work.

\section*{Ethical Impact}
We are committed to upholding intellectual property rights and adhering to all applicable laws and regulations. The images and text instructions included in our benchmark are sourced from publicly available materials, and we have implemented rigorous measures to ensure that no personally sensitive information is present. Furthermore, our efforts are solely dedicated to research purposes and are not intended for commercial use.

\section*{Acknowledgments}
This work was supported by the National Natural Science Foundation of China (62222609, 62320106010), National Science and Technology Major Project (2022ZD0116500), CAS Project for Young Scientists in Basic Research (YSBR-083), and Key Research Program of Frontier Sciences of CAS (ZDBS-LY-7004), Unveiling and Leading Projects of Xiamen (3502Z20241011) , Major Science and Technology Plan Project on the Future Industry Fields of Xiamen City (3502Z20241027) and the InnoHK program. 

\bibliography{custom}

\appendix

\section{Details of the comparison method}
\label{sec:comparison}

In this section, we outline the principles of the baseline methods used in our experiments:

\noindent
\textbf{LwF} mitigates catastrophic forgetting by using knowledge distillation loss during training. Specifically, LwF processes current task inputs through both the current and old models, using the old model’s outputs as "soft labels" to constrain learning via knowledge distillation loss, preserving past knowledge while training on new tasks.

\noindent
\textbf{EWC} mitigates catastrophic forgetting by restricting updates to important weights. It computes parameter importance via the Fisher information matrix and penalizes significant changes, preserving knowledge from previous tasks.

\noindent
\textbf{L2P} introduces a dynamic prompts pool, allowing the model to select and optimize specific prompts during training. Besides, a regularization loss is proposed to  encourage task-specific prompt selection, thereby reducing catastrophic forgetting.

\noindent
\textbf{O-LoRA} enforces an orthogonality loss in parameter space to encourage the current task to optimize in a direction orthogonal to previous tasks, thereby reducing conflicts of different tasks. During inference, it integrates learned knowledge by concatenating the LoRAs of all tasks along the dimension.

\noindent
\textbf{MoELoRA} transforms a single LoRA into a Mixture-of-Experts~(MoE) model with multiple LoRAs based on the number of tasks and trains a router to dynamically assign the appropriate LoRA for each task.

In our experiments, we standardize all methods to the LoRA fine-tuning framework and carefully optimized their hyperparameters to ensure fair and optimal comparisons.

\begin{figure}[H]
    \centering
    \includegraphics[width=0.9\linewidth]{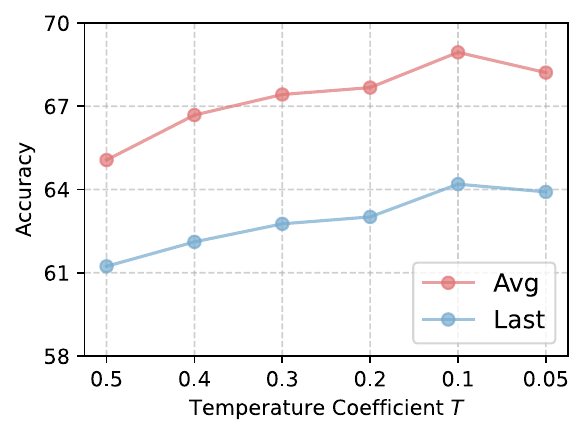}
    \vspace{-10pt}
    \caption{Ablation study of temperature coefficient.}
    \vspace{-10pt}
    \label{fig:temp}
\end{figure}

\section{Details of CKA similarity.}
\label{sec:cka_cal}
The CKA similarity computation process in Section~\ref{CKA} can be summarized as:

(1)~Assuming that $X_i\in \mathbb{R}^{n\times p}$ and $Y_i\in \mathbb{R}^{n\times p}$ represent the output features of two different task inputs at layer $i$, where $n$ is the number of samples and $p$ denotes the dimension of the feature. The first step is to compute the linear kernel matrix: $K_{X_i}=\hat{X_i}\hat{X_i}^T,~K_{Y_i}=\hat{Y_i}\hat{Y_i}^T,$ where $\hat{X_i}, \hat{Y_i}$ denotes the feature after decentralization.

(2)~Next, compute the Hilbert-Schmidt Independence Criterion (HSIC) using linear kernel matrices: $\text{HSIC}(K_{X_i}, K_{Y_i})=\frac{1}{(n-1)^2} \text{tr}(K_{X_i}K_{Y_i})$, where $\text{tr}(\cdot)$ denotes the trace of a matrix.

(3)~Finally, calculate the CKA similarity by normalizing the HSIC values: $\text{CKA}(X_i, Y_i) = \frac{\text{HSIC}(K_{X_i}, K_{Y_i})}{\sqrt{\text{HSIC}(K_{X_i}, K_{X_i})~\text{HSIC}(K_{Y_i}, K_{Y_i})}}.$ 

The value of CKA similarity ranges from 0 to 1, where a higher value indicates greater similarity between the two feature representations. In our work, we compute and compare the CKA similarity for each layer individually using inputs from different datasets.

\section{Ablation study of hyper-parameters}
\label{sec:hyper-parameters}

We perform ablation experiments on the temperature coefficient $T$ in Eq~(\ref{eq:confidence}) in this section. Specifically, we evaluate the impact of different temperature coefficient $T$ on the Avg and Last metrics in the UCIT benchmark. As shown in Fig.~\ref{fig:temp}, smaller temperature coefficients lead to sharper distributions of regularization scores $d_c$, increasing the proportion of LoRA outputs corresponding to the target task and improving overall performance. The chosen value of 0.1 achieves the best overall performance in our experiments.

\begin{table*}[t]
	\centering
         \resizebox{0.9\linewidth}{!}{
	\begin{tabular}{lcc| >{\centering\arraybackslash}p{10.0cm}}
		\toprule	\textbf{Dataset} & \textbf{Train} & \textbf{Test} &  \textbf{Task-specific Instruction} \\
		\midrule \midrule
            ArxivQA & 40000 & 3000 & Answer with the option's letter from the given choices directly. \\
            ImageNet-R & 23998 & 3000 & Answer the question using a single word or phrase. \\
            VizWiz-cap & 40000 & 3000 & Generate a brief caption for the image. \\
            IconQA & 29859 & 3000 & Answer with the option's letter from the given choices directly. \\
            CLEVR-Math & 40000 & 3000 & Answer the question using a single word or phrase. \\
            Flickr30k & 40000 & 3000 & Generate a brief caption for the image. \\
		\bottomrule
	\end{tabular}}
        \vspace{-5pt}
	\caption{Details of datasets used in UCIT benchmark.}
	\label{tab:ucit}
\end{table*}

\begin{table*}[t]
	\centering
         \resizebox{0.9\linewidth}{!}{
	\begin{tabular}{lcc| >{\centering\arraybackslash}p{10.0cm}}
		\toprule	\textbf{Dataset} & \textbf{Train} & \textbf{Test} &  \textbf{Task-specific Instruction} \\
		\midrule \midrule
            GQA & 20000 & 3000 & Answer the question using a single word or phrase. \\
            ImageNet & 20000 & 3000 & Answer the question using a single word or phrase. \\
            TextVQA & 34602 & 3000 & Answer the question using a single word or phrase. \\
            OCRVQA & 20000 & 3000 & Answer the question using a single word or phrase. \\
            REC-COCO & 20000 & 3000 & Please provide the bounding box coordinate of the region this sentence describes: \\
            VizWiz & 20523 & 3000 & Answer the question using a single word or phrase. \\
            VQAv2 & 20000 & 3000 & Answer the question using a single word or phrase. \\
            ScienceQA & 20000 & 3000 & Answer with the option's letter from the given choices directly. \\
		\bottomrule
	\end{tabular}}
        \vspace{-5pt}
	\caption{Details of datasets used in CoIN benchmark.}
	\label{tab:coin}
\end{table*}

\section{Further Experiments}
\label{app:further}
\textbf{Comparison with Model Tailor.} Model Tailor~\cite{zhu2024model} addresses catastrophic forgetting of generic capabilities by identifying and enhancing critical model parameters during downstream task adaptation. However, unlike our setting, which considers continual learning across multiple tasks, Model Tailor only focuses on mitigating forgetting after single-task fine-tuning. To ensure a fair comparison, we preserve Model Tailor's original methodology while evaluating it on our UCIT benchmark under multi-task continual learning conditions. The comparative results are shown in Table~\ref{tab:model_tailor}. While Model Tailor excels on the first task, its performance declines markedly on subsequent tasks, likely due to its need for repeated parameter recalibration. Our method overcomes this limitation through robust knowledge fusion and expansion, maintaining stable performance across tasks.

\begin{table}[H]
    \centering
    \resizebox{0.99\linewidth}{!}{
    \begin{tabular}{clccccccc}
        \toprule
           & {\hspace{1em}} Method & \texttt{\textbf{R}} & \texttt{\textbf{A}} & \texttt{\textbf{V}} & \texttt{\textbf{I}} & \texttt{\textbf{C}} & \texttt{\textbf{F}} & \texttt{{\textbf{Avg}}} \\
        \midrule
            \multirow{2}{*}{\rotatebox{90}{\textbf{Avg}}} 
            & {\hspace{1em}}Model Tailor  & \textbf{89.90} & 71.30 & 15.66 & 39.34 & 29.50 & 44.63 & 48.39  \\
            & {\hspace{1em}}\textbf{HiDe-LLaVA} &{85.70}&\textbf{92.70}&\textbf{54.10}&\textbf{66.87}&\textbf{59.12}&\textbf{55.15}&\textbf{68.94}\\
            \midrule
            % ----------------------------------------------------
            \multirow{2}{*}{\rotatebox{90}{\textbf{Last}}} 
            & {\hspace{1em}}Model Tailor  & \textbf{88.37} & 69.00 & 22.79 & 38.27 & 29.60 & 44.63 & 48.78  \\
            & {\hspace{1em}}\textbf{HiDe-LLaVA}& 80.50&\textbf{89.83}&\textbf{48.78}&\textbf{62.90}&\textbf{47.97}&\textbf{55.15}&\textbf{64.19}\\
        \bottomrule
    \end{tabular}}
    \vspace{-5pt}
    \caption{Comparison with Model Tailor on the UCIT benchmark. \texttt{\textbf{R}:}~\texttt{ImageNet-\textbf{R}}, \texttt{\textbf{A}:}~\texttt{\textbf{A}rxivQA}, \texttt{\textbf{V}:}~\texttt{\textbf{V}izWiz}, \texttt{\textbf{I}:}~\texttt{\textbf{I}conqa}, \texttt{\textbf{C}:}~\texttt{\textbf{C}LEVR-Math}, \texttt{\textbf{F}:}~\texttt{\textbf{F}lickr30k}.}
    \vspace{-5pt}
    \label{tab:model_tailor}
\end{table}

\begin{figure}[t]
    \centering
    \includegraphics[width=0.9\linewidth]{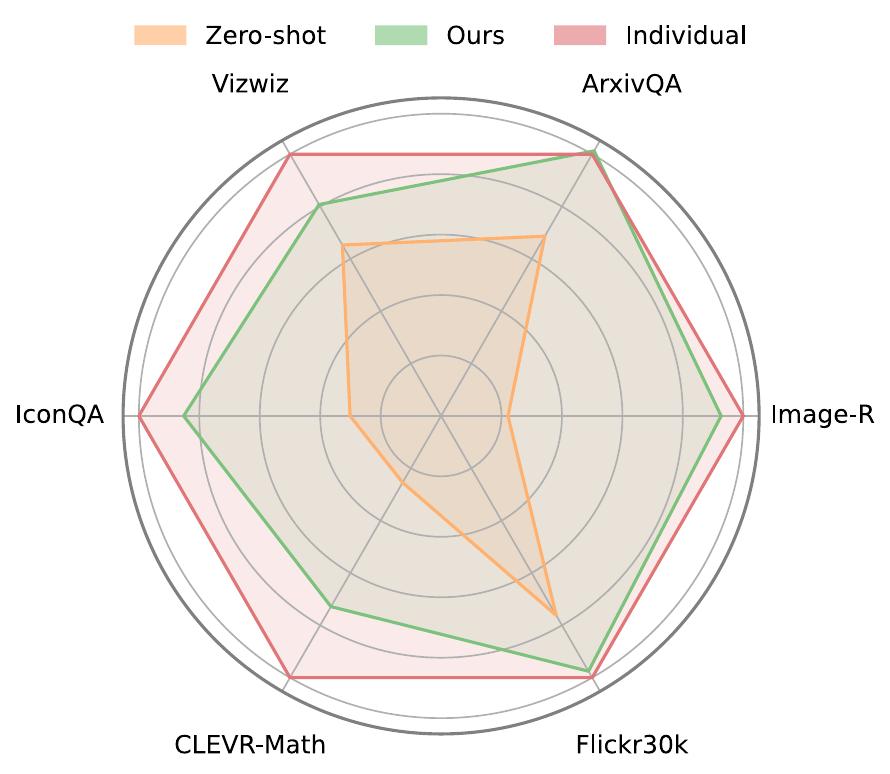}
    \caption{Experiment on InternVL-Chat-7B.}
    \vspace{-15pt}
    \label{fig:radar}
\end{figure}

\noindent
\textbf{Experiment on other MLLM backbone.} To further evaluate the scalability of our proposed method, we conduct experiments on different MLLM architectures. Specifically, we adopt InternVL-Chat-7B~\cite{chen2024internvl} as our base model, which integrates a 6B-parameter Vision Transformer (ViT) as the visual encoder with a 7B-parameter LLM~(\emph{i.e.} Vicuna) via a multimodal projection layer. The experimental results, presented in Fig.~\ref{fig:radar}, demonstrate consistent performance improvements, confirming our method's architecture-agnostic scalability.

\section{Details of UCIT benchmark}
\label{app:detail}

In this paper, we evaluate our method on the CoIN benchmark and the proposed UCIT benchmark. The training and testing scales of these benchmarks, along with the task-specific instruction samples, are detailed in Table~\ref{tab:ucit} and Table~\ref{tab:coin}.

It is important to note that our proposed UCIT benchmark comprises six instruction fine-tuning datasets, selected based on LLaVA's zero-shot performance, which were not included in LLaVA's pre-training phase. This ensures the avoidance of information leakage, a common issue in CoIN. Additionally, we introduce image description tasks to diversify the input instruction types, further enriching the problem formulation.

\section{Visualization of UCIT benchmark}
\label{app:visualization}
As shown in Table~\ref{tab:visualization}, we present examples from various tasks in the UCIT benchmark, where the input domains of most of these datasets differ significantly from LLaVA's original training data domain. Regarding performance, LLaVA's zero-shot performance on several of our proposed datasets is generally low, making them well-suited as benchmarks for continual instruction tuning tasks. In contrast, the datasets proposed by CoIN exhibit substantial overlap with LLaVA's original training data, which fails to effectively capture the catastrophic forgetting across different methods.

\section{Details of evaluation}
In our benchmark, the output forms of tasks are different, so it is necessary to design appropriate way to calculate accuracy for each task. Specifically, for tasks that answer a single option or word, we use \texttt{pred.upper() in ground\_truth.upper()} to determine whether the answer is correct. For caption sentences, we adopt metrics commonly used in image caption tasks to measure the accuracy of the response. Specifically, we use \texttt{Bleu\_1}, \texttt{Bleu\_2}, \texttt{Bleu\_3}, \texttt{Bleu\_4}, \texttt{METEOR}, \texttt{ROUGE\_L}, and \texttt{CIDEr} to comprehensively evaluate the consistency between the ground truth and the responses. For simplicity, we report the average of the seven metrics.

\begin{table*}
    \begin{minipage}{0.99\textwidth}
    \renewcommand\arraystretch{1.0}
    \centering
    \scalebox{0.9}{
    \begin{tabular}{>{\centering\arraybackslash}p{8.0cm} | >{\arraybackslash}p{6.0cm}}
    \toprule
    \vfill \includegraphics[height=3cm, width=3cm]{figure/iconqa_2.pdf} \vfill & 
    \vfill \textbf{Question:} What time is shown? Answer by typing a time word, not a number. It is (\_) to ten. \newline \newline \newline \textbf{Answer:} quarter \vfill \\
    \midrule
    \vfill \includegraphics[height=3cm, width=4cm]{figure/ImgNetR_2.pdf} \vfill & 
    \vfill \textbf{Question:} What is the object in the image? answer the question using a single word or phrase. \newline \newline \newline \textbf{Answer:} Binoculars \vfill \\
    \midrule
    \vfill \includegraphics[height=3cm, width=2cm]{figure/VizWiz_1.pdf} \vfill & 
    \vfill \textbf{Question:} What is happening in the image? Generate a brief caption for the image. \newline \newline \textbf{Answer:} A green and white plastic condiment bottle containing Basil leaves. \vfill \\
    \midrule
    \vfill \includegraphics[height=3cm, width=8cm]{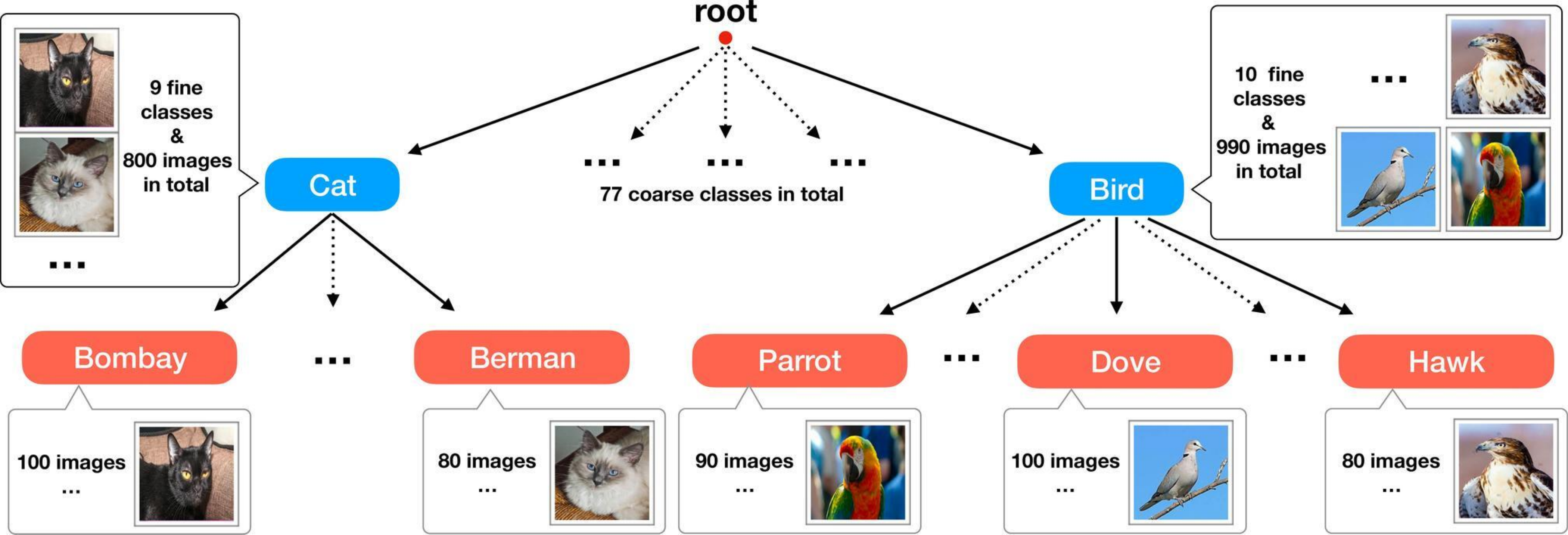} \vfill & 
    \vfill \textbf{Question:} How many coarse classes are represented in the figure? \newline A) Less than 50 \quad ~B) Exactly 77 \newline C) More than 100 ~ D) Exactly 99 \newline Answer with the option's letter from the given choices directly. \newline \textbf{Answer:} B \vfill \\
    \midrule
    \vfill \includegraphics[height=3cm, width=4cm]{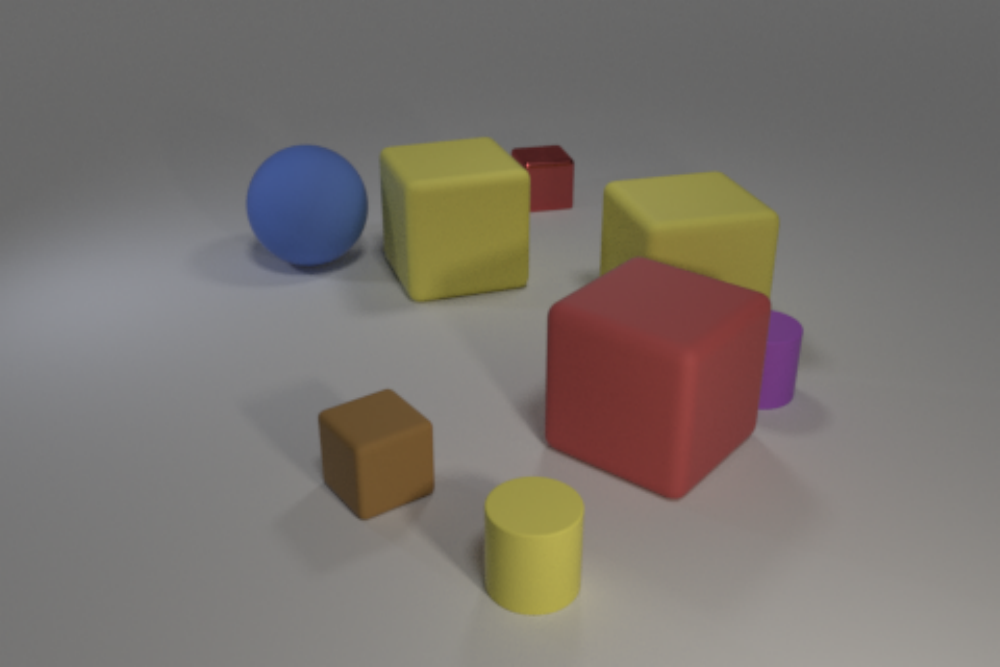} \vfill & 
    \vfill \textbf{Question:} Subtract all brown matte objects. Subtract all blue cylinders. How many objects are left? Answer the question using a single word or phrase. \newline \newline \textbf{Answer:} quarter \vfill \\
    \midrule
    \vfill \includegraphics[height=3cm, width=2cm]{figure/flickr30k_2.pdf} \vfill & 
    \vfill \textbf{Question:} What is happening in the image? Generate a brief caption for the image. \newline \newline \textbf{Answer:} A man in a suit walks along s large building. \vfill \\
    \bottomrule
    \end{tabular}
    }
    \caption{Task data with images across \emph{IconQA}, \emph{ImageNet-R}, \emph{VizWiz-caption}, \emph{ArxivQA}, \emph{CLEVR-Math} and \emph{Flickr30k}.}
    \label{tab:visualization}  
    \end{minipage}
\end{table*}

\end{document}